\documentclass[12pt]{l4dc2020} 
\usepackage{times}

\usepackage[T1]{fontenc}    
\usepackage{url}            
\usepackage{booktabs}       

\usepackage{amssymb}
\usepackage{amsmath}
\usepackage{amsfonts}
\usepackage{mathtools}

\usepackage{bm}             
\usepackage{nicefrac}       
\usepackage{microtype}      

\usepackage{algorithm}
\usepackage{algorithmic}    

\usepackage{mwe}            
\usepackage{xcolor}

\usepackage{hyperref}       
\hypersetup{
 colorlinks=True,
 linkcolor=blue,
 citecolor=blue,
 urlcolor=blue}

\usepackage[frozencache,cachedir=.]{minted}

\usepackage{graphicx}
\graphicspath{{media/}, {../media}}

\usepackage{multicol}       

\newcommand{\name}{\texttt{Lyceum}}
\newcommand{\namebase}{\texttt{LyceumBase.jl}}

\newcommand{\namemujocojl}{\texttt{LyceumMuJoCo.jl}}
\newcommand{\nameaijl}{\texttt{LyceumAI.jl}}
\newcommand{\namevizjl}{\texttt{LyceumMuJoCoViz.jl}}
\newcommand{\mujocojl}{\texttt{MuJoCo.jl}}
\newcommand{\lyceum}{\name}

\title{Lyceum: An efficient and scalable ecosystem for robot learning}

\author{
\begin{center}
\Name{Colin Summers}$^1$, 
\Name{Kendall Lowrey}$^1$, 
\Name{Aravind Rajeswaran}$^1$ \\[2pt]
\Name{Siddhartha Srinivasa}$^1$,
\Name{Emanuel Todorov}$^{1,2}$ \\[5pt]
{\tt \{colinxs, klowrey, aravraj, siddh, todorov\}@cs.uw.edu} \\[2pt]
\addr $^1$ University of Washington Seattle, $^2$ Roboti LLC
\end{center}
}

\begin{document}

\maketitle

\vspace*{-1cm}
\begin{abstract}
    We introduce~\lyceum, a high-performance computational ecosystem for robot learning. \name~is built on top of the Julia programming language and the MuJoCo physics simulator, combining the ease-of-use of a high-level programming language with the performance of native C. In addition, \name~has a straightforward API to support parallel computation across multiple cores and machines. Overall, depending on the complexity of the environment, \name~is 5-30X faster compared to other popular abstractions like OpenAI's \texttt{Gym} and DeepMind's \texttt{dm-control}. This substantially reduces training time for various reinforcement learning algorithms; and is also fast enough to support real-time model predictive control through MuJoCo.
    The code, tutorials, and demonstration videos can be found at: \url{www.lyceum.ml}.
\end{abstract}

\section{Introduction} \label{sec:intro}
Progress in artificial intelligence has exploded in recent years, due in large part to advances computational infrastructure. The advent of massively parallel GPU computing, combined with powerful automatic-differentiation tools like TensorFlow~\citep{tensorflow} and PyTorch~\citep{pytorch}, have lead to new classes of deep learning algorithms by enabling what was once computationally intractable. These tools, alongside fast and accurate physics simulators like MuJoCo~\citep{mujoco12}, and associated frameworks like OpenAI's Gym~\citep{gym} and DeepMind's dm\_control~\citep{DeepmindDmControl2019}, have similarly transformed various aspects of robotic control like Reinforcement Learning (RL), Model-Predictive Control (MPC), and motion planning. These platforms enable researchers to give their ideas computational form, share results with collaborators, and deploy their successes on real systems. 

From these advances, simulation to real-world~\texttt{(sim2real)} transfer has emerged as a promising paradigm for robotic control. A growing body of recent work suggests that robust control policies trained in simulation can successfully transfer to the real world~\citep{OpenAI2018LearningDI, Rajeswaran2016EPOpt, Sadeghi2016CAD2RLRS, Lowrey2018ReinforcementLF, Tobin2017DRGrasping, Mordatch2015EnsembleCIO}. 
However, many algorithms used in these works for controller synthesis are computationally intensive. 
Training control policies with state-of-the-art RL algorithms often takes many hours to days of compute time. For example, OpenAI's landmark Dactyl work~\citep{OpenAI2018LearningDI} required 50 hours of training time across 6144 CPU cores and 8 NVIDIA V100 GPUs. 
Such computational budgets are available only to a select few labs. Furthermore, such experiments are seldom run only once in deep learning and especially in deep RL. Indeed, RL algorithms are notoriously sensitive to choices of hyper-parameters~\citep{Rajeswaran17nips, Henderson2017DeepRL, Mania2018SimpleRS}. Thus, many iterations of the learning process may be required, with humans in the loop, to improve hyperparameter choices and reward functions, before finally deploying solutions in the real world. This computational bottleneck often leads to a scarcity of hardware results, relative to the number of papers that propose new algorithms on highly simplified and well tuned benchmark tasks. Exploring avenues to reduce experiment turn around time is thus crucial for scaling up to harder tasks and making resource-intensive algorithms and environments accessible to research labs without massive cloud computing budgets.

In a similar vein, computational considerations have also limited progress in model-based control algorithms. For real-time model predictive control~(MPC), the computational restrictions manifest as the requirement to compute actions in bounded time with limited local resources. As we will show, existing frameworks such as Gym and dm\_control, while providing a convenient abstraction in Python, are too slow to meet this real-time computation requirement. As a result, most planning algorithms are run offline and deployed in open-loop mode on hardware. This is unfortunate, since it does not take feedback into account which is well known to be critical for stochastic control.

\paragraph{Our Contributions:}
Our goal in this work is to overcome the aforementioned computational restrictions to enable faster training of policies with RL algorithms, facilitate real-time MPC with a detailed physics simulator, and ultimately enable researchers to engage with complex robotic tasks. To this end, we develop \name, a computational ecosystem that uses the Julia programming language and the MuJoCo physics engine. \name~ships with the main OpenAI gym continuous control tasks, along with other environments representative of challenges in robotics. Julia's unique features allow us to wrap MuJoCo with zero-cost abstractions, providing the flexibility of a high-level programming language to enable easy creation of environments, tasks, and algorithms, while retaining the performance of native C. This allows RL and MPC algorithms implemented in \name~to be \hbox{5-30X} faster compared to Gym and dm\_control. We hope that this speedup will enable RL researchers to scale up to harder problems with reduced computational costs, and also enable real-time MPC.

\section{Related Works}
Recently, various physics simulators and the computational ecosystems surrounding them have transformed robot learning research. They allow for exercising creativity to quickly generate new and interesting robotic scenes, as well as quickly prototype various learning and control solutions. We summarize the main threads of related work below.

\paragraph{Physics simulators:} MuJoCo~\citep{mujoco12} has quickly emerged as a leading physics simulator for robot learning research. It is fast and efficient, and particularly well suited for contact-rich tasks. Numerous recent works have also demonstrated simulation to reality transfer with MuJoCo through physically consistent system identification~\citep{Lowrey2018ReinforcementLF} or domain randomization~\citep{OpenAI2018LearningDI, Mordatch2015EnsembleCIO, Nachum2019QuadrupedSim2Real}. Our framework wraps MuJoCo in Julia and enables programming and research with a high level language, while retaining the speed of native C. While we primarily focus on MuJoCo, we believe that similar design principles can be extended to other simulators like Bullet~\citep{Bullet} and DART~\citep{DARTSimulator}.

\paragraph{Computational ecosystems:} 
OpenAI's gym~\citep{gym} and DeepMind's dm\_control \citep{DeepmindDmControl2019} sparked a wave of interest by providing Python bindings for MuJoCo with a high-level API, as well as easy-to-use environments and algorithms. This has enabled the RL community to quickly access physics-based environments and prototype algorithms. Unfortunately, this flexibility comes at the price of computational efficiency. Existing ecosystems are slow due to inefficiencies and poor parallelization capabilities of Python. Prior works have tried to address some of the shortcomings of Python-based frameworks by attempting to add JIT compilation to the language~\citep{lamNumbaLLVMbasedPython2015, pytorch, agrawalTensorFlowEagerMultiStage2019} but only support a subset of the language, and do not achieve the same performance as Julia. \citet{surreal} developed a framework similar to Gym that supports distributed computing, but it still suffers the same performance issues of Python and multi-processing. Perhaps closest to our motivation is the work of~\citet{Koolen2019JuliaFR}, which demonstrates the usefulness of Julia as a language for robotics. However, it uses a custom and minimalist rigid body simulator with limited contact support. In contrast, our work addresses the inefficiencies of existing computational ecosystems through use the of Julia, and directly wraps a more capable simulator, MuJoCo, with zero overhead.

\paragraph{Algorithmic toolkits and environments:} A number of algorithmic toolkits like OpenAI Baselines~\citep{baselines}, mjRL~\citep{Rajeswaran17nips}, Soft-Learning~\citep{softlearning}, and RL-lab~\citep{rllab}; as well as environments like Hand Manipulation Suite~\citep{Rajeswaran-RSS-18}, Robel~\citep{Kumar_ROBEL}, Door Gym~\citep{Urakami2019DoorGymAS}, and Surreal Robosuite~\citep{surreal} have been developed around existing computational ecosystems. Our framework supports all the underlying functionality needed to transfer these advances into our ecosystem (e.g. simulator wrappers and automatic differentiation through Flux). \name~ comes with a few popular algorithms out of the box like NPG~\citep{kakade2002natural, Rajeswaran17nips} for RL and variants of MPPI~\citep{POLO, williams2016aggressive} for MPC. In the future, we plan to port further algorithms and advances into our ecosystem and look forward to community contributions as well.

\section{The \name~ecosystem}
The computational considerations are unique for designing infrastructure and ecosystems for robotic control with RL and MPC. We desire a computational ecosystem that is high-level and easy to use for research, can efficiently handle parallel operations, while ideally also matching serial operation speed of native C to be usable on robots. We found Julia to be well suited for these requirements, and we summarize some of these main advantages that prompted us to use Julia. Subsequently, we outline some salient features of \name.

\subsection{Julia for robotics and RL}

Julia is a general-purpose programming language developed in 2012 at MIT with a focus on technical computing \citep{JuliaMicrobenchmarks2019}. While a full description of Julia is beyond the scope of this paper, we \textbf{}highlight a few key aspects that we leverage in \name~and believe make Julia an excellent tool for robotics and RL researchers.

\paragraph{Just-in-time compilation}
Julia feels like a dynamic, interpreted scripting language, enabling an interactive programming experience.
Under the hood, however, Julia leverages the LLVM backend to "just-in-time"~(JIT) compile native machine code that is as fast as C for a variety of hardware platforms \citep{JuliaMicrobenchmarks2019}. This enables researchers to quickly prototype ideas and optimize for performance with the same language.

\paragraph{Julia can easily call functions in Python and C}
In addition to the current ecosystem of Julia packages, users can interact with Python and C as illustrated below. 
This allows researchers to benefit from existing body of deep learning research (in Python), and also interact easily with low-level robot hardware drivers.
\begin{minted}[fontsize=\footnotesize]{julia}
  using PyCall
  so = pyimport("scipy.optimize")
  so.newton(x -> cos(x) - x, 1)
  ccall((:mjr_getError,libmujoco),Cint,())
\end{minted}

\paragraph{Easy paralellization}
Julia comes with extensive support for distributed and shared-memory multi-threading that allows users to trivially parallelize their code. The following example splits the indices of $X$ across all the available cores and performs in-place multiplication in parallel:
\begin{minted}[fontsize=\footnotesize]{julia}
  @threads for i in eachindex(X)
    X[i] *= 2
  end
\end{minted}
Julia can also transpile to alternative hardware backends, allowing use of parallel processors like GPUs by writing high level Julia code.

\paragraph{Simple package management}
To handle the 3000+ packages available, Julia comes with a built-in package manager, avoiding "dependency hell", and facilitating collaboration and replication. This means less time is spent getting things to run and more time for focusing on the task at hand.

\subsection{Salient Features of \name}

\name~consists of the following packages 
\begin{enumerate}
\itemsep-0.25em
    \item \namebase, a "base" package which defines a set of abstract environment and controller interfaces, along with several utilities.
    \item \mujocojl, a low-level Julia wrapper for the MuJoCo physics simulator.
    \item \namemujocojl, a high-level "environment" abstraction similar to Gym and dm\_control.
    \item \namevizjl, a flexible policy and trajectory visualizer with interaction.
    \item \nameaijl, a collection of various algorithms for robotic control.
\end{enumerate}

\paragraph{\namebase}
At the highest level we provide \namebase, which contains several convenience utilities used throughout the \name~ecosystem for data logging, multi-threading, and controller benchmarking (i.e. measuring throughput, jitter, etc.). \namebase~also contains interface definitions, such as \mintinline{julia}{AbstractEnvironment} which \namemujocojl~implements. See the appendix for the full \mintinline{julia}{AbstractEnvironment} interface. 

This interface is similar to the popular Python frameworks Gym and dm\_control, where an agent's observations are defined, actions are chosen, and the simulator can step. A few key differences are as follows:
\begin{enumerate}
\itemsep-0.25em
    \item The ability to arbitrarily get/set the state of the simulator, a necessary feature for model-based methods like MPC or motion planning. An important component of this is defining a proper notion of a state, which is often missing from existing frameworks, as it can include more than just the position and velocities of the dynamics.
    \item Optional, in-place versions for all functions (e.g. getstate!$(\cdot)$) which store the return value in a pre-allocated data structure. This eliminates unnecessary memory allocations and garbage collection, enabling environments to be used in tight, real-time control loops.
    \item An optional "evaluation" metric. Often times reward functions are heavily "shaped" and hard to interpret. For example, the reward function for bipedal walking may include root pose, ZMP terms, control costs, etc., while success can instead be simply evaluated by distance of the root along an axis.
\end{enumerate}
We expect most users will be interested in implementing their own environments, which forms a crucial part of robotics research. Indeed, different researchers may be interested in different robots performing different tasks, ranging from whole arm manipulators to legged locomotion to dexterous anthropomorphic hands. To aid this process, we provide sensible defaults for most of the API, making it easy to get started and experiment with different environments.
The separation of interface and implementation also allows for other simulators and back-ends (e.g. RigidBodySim.jl or DART) to be used in lieu of the MuJoCo-based environments we provide, should the user desire.

\paragraph{\mujocojl, \namemujocojl, and \namevizjl}
\mujocojl~is a low-level Julia wrapper for MuJoCo that has a one-to-one correspondence to MuJoCo 2.0's C interface and includes soft body dynamics. All data is memory mapped with no overhead, and named fields in a MuJoCo.xml file are exposed to the data structures, enabling field access as \mintinline{julia}{d.qpos[:, :arm]}. We then build \namemujocojl, the MuJoCo implementation of our \mintinline{julia}{AbstractEnvironment} API, on top of \mujocojl. This is the construction of an environment based in MuJoCo, and allows the user to configure tasks rewards and programatically modify dynamics before passing the structure to algorithms for processing. Finally the \namevizjl~package visualizes the results of MuJoCo based models. Data is passed in as a list of trajectories for viewing or control function; a trained policy or MPC controller, for example, can be passed to the visualizer, which has hooks for keyboard and mouse interaction. Robots in the real world encounter perturbations and disturbances, and with \namevizjl~the user can interact with the simulated environment to test the robustness of a controller.

\paragraph{\nameaijl}
Coupled with these environments is \nameaijl, a collection of algorithms for robotic control that similarly leverage Julia's performance and multi-threading abilities. Currently we provide implementations of "Model Predictive Path Integral Control" (MPPI), a stochastic shooting method for model-predictive control and Natural Policy Gradient. We compare these methods with a Python implementation in the next sections. Both of these methods benefit from multi-threaded rollouts, either with respect to controls or a policy, which can be performed in parallel. Neural networks and automatic differentiation for objects like control policies or fitted value functions are handled by Flux.jl and Zygote.jl \citep{Flux.jl-2018,Zygote.jl-2018}, which are also Julia based. The combination of efficient compute utilization, flexible high level programming, and an ecosystem of tools should allow both robotics and RL researchers to experiment with different robotic systems, algorithm design, and hopefully deploy to real systems.

\section{Benchmark Experiments and Results} \label{sec:benchmark_experiments}
\begin{figure}[b!]
    \centering
    \includegraphics[width=0.49\textwidth]{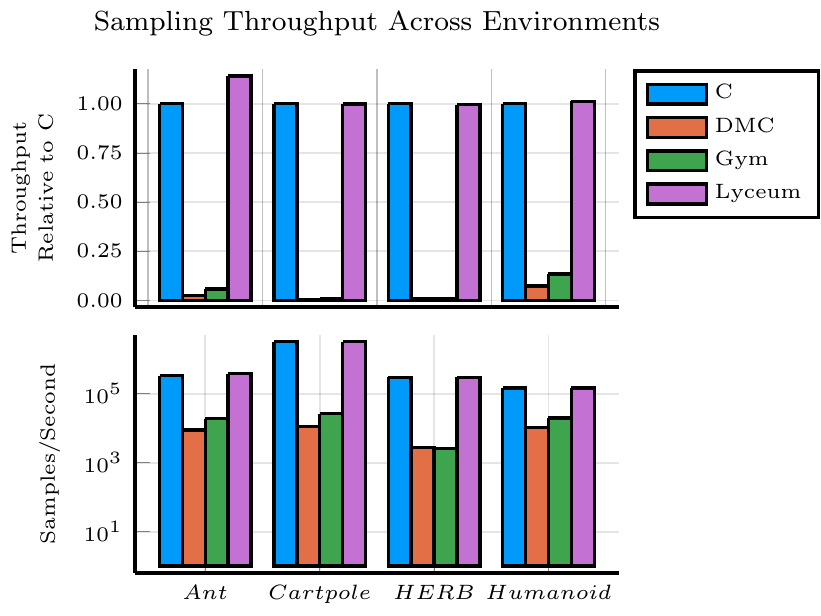}
    \includegraphics[width=0.49\textwidth]{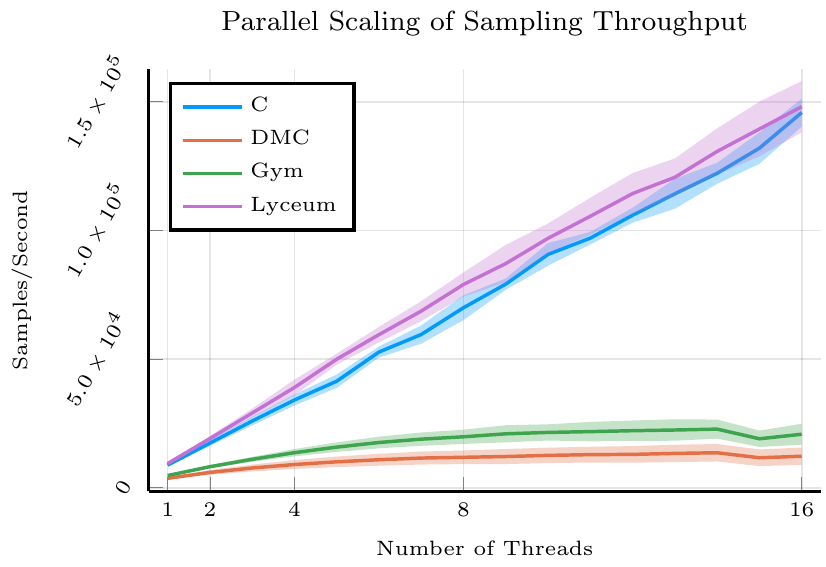}

    \caption{(Left) Comparison of sampling throughput for models of varying complexity. In the top left plot, we provide the sampling throughput relative to native C, while in the bottom left plot, we provide samples per second. This experiment uses all 16 available cores (see main text for details). We find that \name~can match the performance of native C, while still retaining the benefits of writing research code in a high-level language. (Right) Sampling throughput as a function of threads for the Humanoid model.}
    \label{fig:env_throughput}
\end{figure}

We designed our experiments and timing benchmarks to answer the following questions: (a) Do the implementations of Gym RL environments and algorithms in \name~produce comparable results? (b) Does \name~lead to faster environment sampling and experiment turn-around time when compared to Gym and dm\_control?

\paragraph{Experiment Setup}
All experiments are performed on a 16-core Intel i9-7960X with the CPU governor pinned at 1.2GHz so as to prevent dynamic scaling or thermal throttling from affecting results. As Gym and dm\_control do not come with built-in support for parallel computing, the authors implement this functionality using Python's \mintinline{python}{multiprocessing} library as recommended in several GitHub Issues by the respective library authors. Below we describe the various benchmarks we considered and their results.

\paragraph{Sampling Throughput} In the first benchmark, we study the sampling throughput and parallel scaling performance of \namemujocojl~against Gym, dm\_control, and a native C implementation using an OpenMP thread pool. To do so, we consider various models of increasing complexity: \texttt{CartPole, Ant, HERB,} and \texttt{Humanoid.} In the first experiment, we use all 16 of the available cores to measure the number of samples we can collect per second. Figure \ref{fig:env_throughput} (left) shows the results, which are presented in two forms: as a fraction of native C's throughput, and as samples per second. We see that \name~and native C significantly outperform Gym and dm\_control in all cases. In particular, for CartPole, \name~is more than 200x faster compared to Gym.

In the second experiment, we study how the sampling performance scales with the number of cores for the various implementations. To do so, we consider the Humanoid model and measure the number of samples that can be generated per second with varying number of cores. The results are presented in Figure~\ref{fig:env_throughput} (right), where we see substantial gains for \name. In particular, the scaling is near linear with the number of cores for C and \name, while there are diminishing returns for Gym and dm\_control. This is due to inherent multi-threading limitations of (pure) Python. When using more cores (e.g. on a cluster), the performance difference is likely to be even larger.

\begin{figure}[t!]
    \centering
    \includegraphics{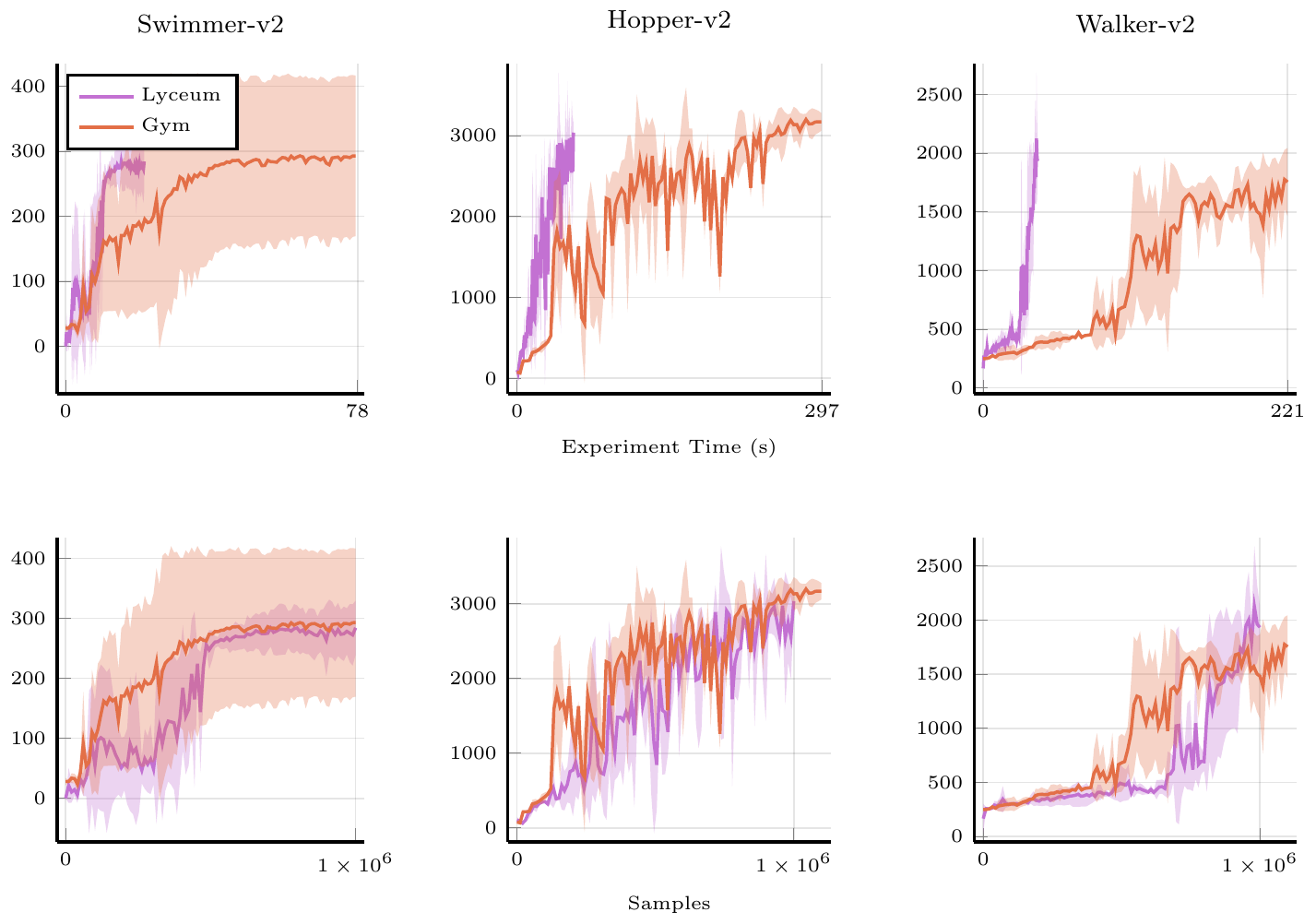}
    \caption{Reinforcement learning with NPG in the Gym and \name~frameworks, training for one million time-steps. Top row presents environment reward vs experiment (i.e. wall clock time), and bottom row presents environment reward vs number of simulated timesteps. Performance of the underlying deterministic policy is reported.}
    \label{fig:npg_gym_comparison}
\end{figure}

\begin{figure}
    \begin{minipage}{\textwidth}
    \begin{minipage}[t!]{0.285\textwidth}
    \centering
    \includegraphics[width=0.95\linewidth]{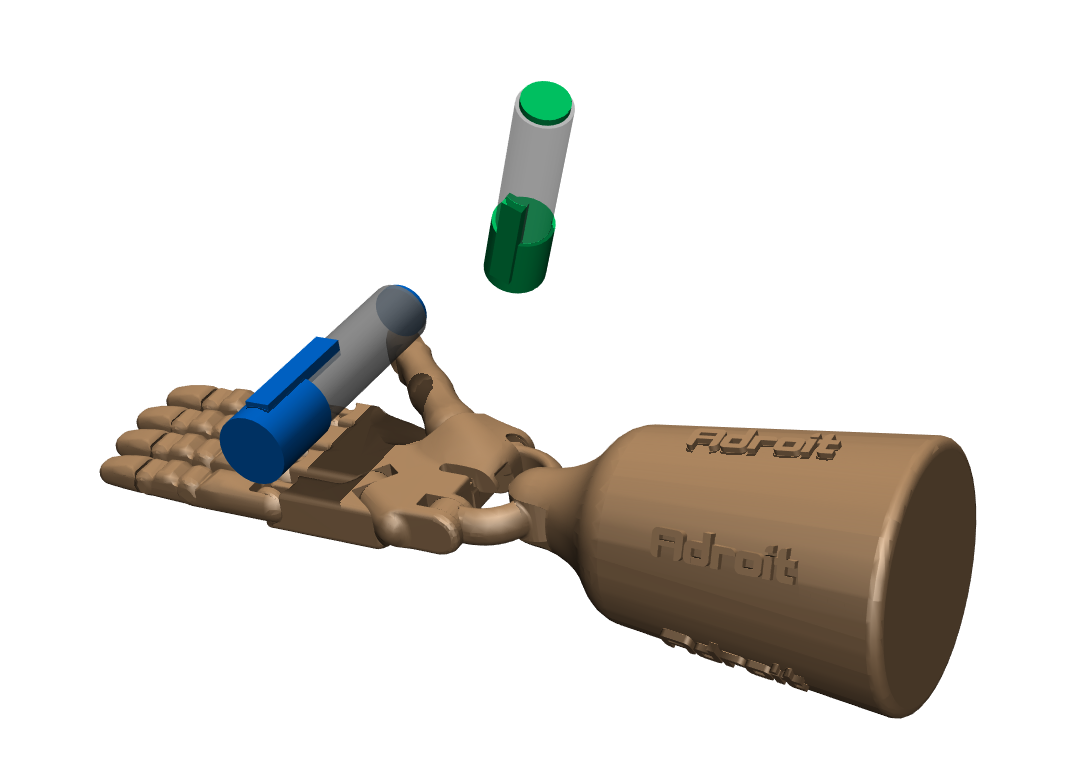}
    \end{minipage}
    \begin{minipage}[t!]{0.205\textwidth}
    \centering
    \includegraphics[width=0.95\linewidth]{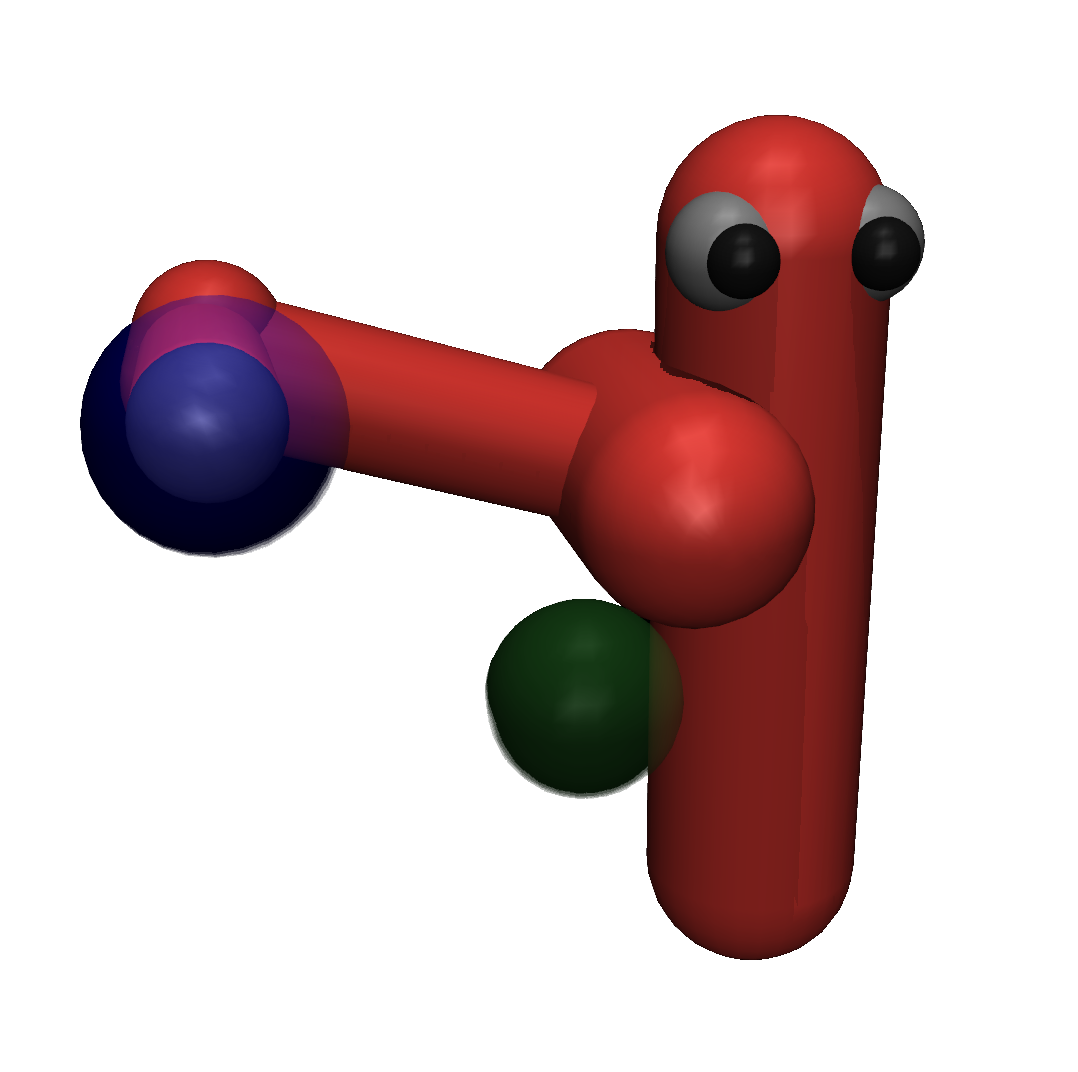}
    \end{minipage}
    \hfill
    \begin{minipage}[t!]{0.48\textwidth}
    \begin{tabular}[t!]{l|c|c}
        Task                & Time (s) & Success (\%) \\ \hline
        Hand~(gym)          & 40.4               &  92 $\pm$ 5.31 \% \\
        Hand~(ours)         & \textbf{14.3}      &  88 $\pm$ 6.36 \% \\
        Reacher~(gym)       & 3.67               &  100\% \\
        Reacher~(ours)      & \textbf{0.119}     &  100\%  \\
    \end{tabular}
    \end{minipage}
    \caption{\label{fig:mppi_comparison} (Left) Illustration of the 30-DOF in-hand manipulation task with a Shadow Hand~(Adroit). The goal is to manipulate the (blue) pen to match the (green) desired pose. (Middle) Illustration of the reaching task with a 7-DOF Sawyer arm. Goal is to make the end-effector (blue) reach the (green) target. (Right)~comparison of time taken and success percentage in gym and \name. Time refers to the time taken to execute a single episode with the MPPI controller (in MPC mode). Success \% measures the number of successful episodes when the robot is controlled using the MPPI algorithm. $95\%$ confidence intervals are also reported. See appendix for additional details and hyperparameters.
    }
  \end{minipage}
\end{figure}

\paragraph{Reinforcement Learning with Policy Gradients} In the second benchmark, we compare the learning curves and wall clock time of Natural Policy Gradient~\citep{kakade2002natural, Rajeswaran17nips}, which is closely related to TRPO~\citep{Schulman15}, between Gym and \name. Our implementation of NPG, closely based on the algorithm as described in \citet{Rajeswaran17nips}
and consistent with majority practice in the community, considers 2 layer neural network policies. Details about hyperparameters are provided in the appendix. We compare based on three representative tasks (Swimmer, Hopper, and Walker) and find that the learning curves match across the two frameworks. The results are summarized in Figure~\ref{fig:npg_gym_comparison}, and we find that the performance curves match well. We note that RL algorithms are known to be sensitive to many implementation details~\citep{Henderson2017DeepRL, Ilyas2018AreDP}, and thus even approximately matching results is a promising sign for both the original code base and \name. We expect that with further code-level optimization, the performance of RL algorithms in \name~can match their counterparts in Gym.

\paragraph{Model Predictive Control}
In the final benchmark, we compare the performance of a model-based trajectory optimizer on Gym and \name. For this purpose, we consider the Model Predictive Path Integral~(MPPI) algorithm~\citep{williams2016aggressive}, which in conjunction with learning based techniques have demonstrated impressive results in tasks like aggressive driving and dexterous hand manipulation~\citep{Lowrey2018ReinforcementLF, PDDM}. MPPI is a sampling based algorithm where different candidate action sequences are considered to generate many potential trajectories starting from the current state. Rewards are calculated for each of these trajectories, and the candidate action sequences are combined with exponentially weighted trajectory rewards. 

We consider two tasks for the MPPI comparison: a 7-DOF sawyer arm that reaches various spatial goals with the end effector, and a 30-DOF in-hand manipulation task where a Shadow hand (Adroit)~\citep{Kumar2016thesis} has to perform in-hand manipulation of a pen to match a desired configuration. We compare the times taken by MPPI to optimize a trajectory and also the fraction of times MPPI optimized a successful trajectory. The results are provided in Figure~\ref{fig:mppi_comparison}. In summary, we find that the MPPI success percentage are comparable in both Gym and \name, and the \name~implementation is approx. 30x faster for the Sawyer arm task and 3x faster for the Shadow Hand task. This trend is consistent with the earlier trend, where the relative differences are larger for lower dimensional systems with fewer contacts. This is because for complex models with many contacts like the Shadow Hand, most of the computational work is performed by MuJoCo, thereby diminishing the impact of overheads in Gym. We also note, however, that we found the performance scaling with cores to be better in \name~compared to Gym, and thus the difference between the frameworks is likely larger when using more cores (e.g. on a cluster).

\section{Conclusion and Future Work}
We introduced \name,~a new computational ecosystem for robot learning in Julia that provides the rapid prototyping and ease-of-use benefits of a high-level programming language, yet retaining the performance of a low-level language like C. We demonstrated that this ecosystem can obtain substantial speedups compared to existing ecosystems like OpenAI gym and dm\_control. We also demonstrated that this speed up enables faster experimental times for RL and MPC algorithms. In the future, we hope to port over algorithmic infrastructures like OpenAI's baselines~\citep{baselines}. We also hope to include and support models and environments involving real robots like Cassie, HERB~\citep{Srinivasa2009HERBAH}, ROBEL~\citep{Kumar_ROBEL}, and Shadow Hand~\citep{Kumar2016thesis, Rajeswaran-RSS-18, Jain-ICRA-19}, and also support for fast rendering to enable research integrating perception and control. 

\section*{Acknowledgements}
The authors thank Ben Evans for continued valuable feedback over the course of the project. The authors also thank Vikash Kumar and Svet Kolev for valuable comments about the paper draft, and Motoya Ohnishi for early adoption of Lyceum and feedback. This work was (partially) funded by the National Institute of Health R01 (\#R01EB019335), National Science Foundation CPS (\#1544797), National Science Foundation NRI (\#1637748), the Office of Naval Research, the RCTA, Amazon, and Honda Research Institute USA.

\bibliography{references}

\begin{thebibliography}{39}
\providecommand{\natexlab}[1]{#1}
\providecommand{\url}[1]{\texttt{#1}}
\expandafter\ifx\csname urlstyle\endcsname\relax
  \providecommand{\doi}[1]{doi: #1}\else
  \providecommand{\doi}{doi: \begingroup \urlstyle{rm}\Url}\fi

\bibitem[Abadi et~al.(2016)Abadi, Barham, Chen, Chen, Davis, Dean, Devin,
  Ghemawat, Irving, Isard, Kudlur, Levenberg, Monga, Moore, Murray, Steiner,
  Tucker, Vasudevan, Warden, Wicke, Yu, and Zhang]{tensorflow}
Mart{\'i}n Abadi, Paul Barham, Jianmin Chen, Zhifeng Chen, Andy Davis, Jeffrey
  Dean, Matthieu Devin, Sanjay Ghemawat, Geoffrey Irving, Michael Isard,
  Manjunath Kudlur, Josh Levenberg, Rajat Monga, Sherry Moore, Derek~Gordon
  Murray, Benoit Steiner, Paul~A. Tucker, Vijay Vasudevan, Pete Warden, Martin
  Wicke, Yuan Yu, and Xiaoqiang Zhang.
\newblock Tensorflow: A system for large-scale machine learning.
\newblock In \emph{OSDI}, 2016.

\bibitem[Agrawal et~al.(2019)Agrawal, Modi, Passos, Lavoie, Agarwal, Shankar,
  Ganichev, Levenberg, Hong, Monga, and
  Cai]{agrawalTensorFlowEagerMultiStage2019}
Akshay Agrawal, Akshay~Naresh Modi, Alexandre Passos, Allen Lavoie, Ashish
  Agarwal, Asim Shankar, Igor Ganichev, Josh Levenberg, Mingsheng Hong, Rajat
  Monga, and Shanqing Cai.
\newblock Tensorflow eager: A multi-stage, python-embedded dsl for machine
  learning.
\newblock \emph{ArXiv}, abs/1903.01855, 2019.

\bibitem[Ahn et~al.(2019)Ahn, Zhu, Hartikainen, Ponte, Gupta, Levine, and
  Kumar]{Kumar_ROBEL}
Michael Ahn, Henry Zhu, Kristian Hartikainen, Hugo Ponte, Abhishek Gupta,
  Sergey Levine, and Vikash Kumar.
\newblock {ROBEL: RObotics BEnchmarks for Learning with low-cost robots}.
\newblock In \emph{Conference on Robot Learning (CoRL)}, 2019.

\bibitem[Bezanson et~al.(2017)Bezanson, Edelman, Karpinski, and
  Shah]{JuliaMicrobenchmarks2019}
Jeff Bezanson, Alan Edelman, Stefan Karpinski, and Viral~B Shah.
\newblock Julia: A fresh approach to numerical computing.
\newblock \emph{SIAM review}, 59\penalty0 (1):\penalty0 65--98, 2017.
\newblock URL \url{https://doi.org/10.1137/141000671}.

\bibitem[Brockman et~al.(2016)Brockman, Cheung, Pettersson, Schneider,
  Schulman, Tang, and Zaremba]{gym}
Greg Brockman, Vicki Cheung, Ludwig Pettersson, Jonas Schneider, John Schulman,
  Jie Tang, and Wojciech Zaremba.
\newblock Openai gym, 2016.

\bibitem[Coumans(2013)]{Bullet}
Erwin Coumans.
\newblock Bullet physics library, 2013.

\bibitem[Dhariwal et~al.(2017)Dhariwal, Hesse, Klimov, Nichol, Plappert,
  Radford, Schulman, Sidor, Wu, and Zhokhov]{baselines}
Prafulla Dhariwal, Christopher Hesse, Oleg Klimov, Alex Nichol, Matthias
  Plappert, Alec Radford, John Schulman, Szymon Sidor, Yuhuai Wu, and Peter
  Zhokhov.
\newblock Openai baselines.
\newblock \url{https://github.com/openai/baselines}, 2017.

\bibitem[Duan et~al.(2016)Duan, Chen, Houthooft, Schulman, and Abbeel]{rllab}
Yan Duan, Xi~Chen, Rein Houthooft, John Schulman, and Pieter Abbeel.
\newblock Benchmarking deep reinforcement learning for continuous control.
\newblock \emph{ArXiv}, abs/1604.06778, 2016.

\bibitem[Fan et~al.(2018)Fan, Zhu, Zhu, Liu, Zeng, Gupta, Creus-Costa,
  Savarese, and Fei-Fei]{surreal}
Linxi Fan, Yuke Zhu, Jiren Zhu, Zihua Liu, Orien Zeng, Anchit Gupta, Joan
  Creus-Costa, Silvio Savarese, and Li~Fei-Fei.
\newblock Surreal: Open-source reinforcement learning framework and robot
  manipulation benchmark.
\newblock In \emph{Conference on Robot Learning}, 2018.

\bibitem[Haarnoja et~al.(2018)Haarnoja, Zhou, Hartikainen, Tucker, Ha, Tan,
  Kumar, Zhu, Gupta, Abbeel, and Levine]{softlearning}
Tuomas Haarnoja, Aurick Zhou, Kristian Hartikainen, George Tucker, Sehoon Ha,
  Jie Tan, Vikash Kumar, Henry Zhu, Abhishek Gupta, Pieter Abbeel, and Sergey
  Levine.
\newblock Soft actor-critic algorithms and applications.
\newblock Technical report, 2018.

\bibitem[Henderson et~al.(2017)Henderson, Islam, Bachman, Pineau, Precup, and
  Meger]{Henderson2017DeepRL}
Peter Henderson, Riashat Islam, Philip Bachman, Joelle Pineau, Doina Precup,
  and David Meger.
\newblock Deep reinforcement learning that matters.
\newblock \emph{ArXiv}, abs/1709.06560, 2017.

\bibitem[Ilyas et~al.(2018)Ilyas, Engstrom, Santurkar, Tsipras, Janoos,
  Rudolph, and Madry]{Ilyas2018AreDP}
Andrew Ilyas, Logan Engstrom, Shibani Santurkar, Dimitris Tsipras, Firdaus
  Janoos, Larry Rudolph, and Aleksander Madry.
\newblock Are deep policy gradient algorithms truly policy gradient algorithms?
\newblock \emph{ArXiv}, abs/1811.02553, 2018.

\bibitem[Innes(2018)]{Zygote.jl-2018}
Michael Innes.
\newblock Don't unroll adjoint: Differentiating ssa-form programs.
\newblock \emph{CoRR}, abs/1810.07951, 2018.
\newblock URL \url{http://arxiv.org/abs/1810.07951}.

\bibitem[Innes et~al.(2018)Innes, Saba, Fischer, Gandhi, Rudilosso, Joy,
  Karmali, Pal, and Shah]{Flux.jl-2018}
Michael Innes, Elliot Saba, Keno Fischer, Dhairya Gandhi, Marco~Concetto
  Rudilosso, Neethu~Mariya Joy, Tejan Karmali, Avik Pal, and Viral Shah.
\newblock Fashionable modelling with flux.
\newblock \emph{CoRR}, abs/1811.01457, 2018.
\newblock URL \url{http://arxiv.org/abs/1811.01457}.

\bibitem[Jain et~al.(2019)Jain, Li, Singhal, Rajeswaran, Kumar, and
  Todorov]{Jain-ICRA-19}
Divye Jain, Andrew Li, Shivam Singhal, Aravind Rajeswaran, Vikash Kumar, and
  Emanuel Todorov.
\newblock {Learning Deep Visuomotor Policies for Dexterous Hand Manipulation}.
\newblock In \emph{International Conference on Robotics and Automation (ICRA)},
  2019.

\bibitem[Kakade(2002)]{kakade2002natural}
Sham~M Kakade.
\newblock A natural policy gradient.
\newblock In \emph{NIPS}, 2002.

\bibitem[Koolen and Deits(2019)]{Koolen2019JuliaFR}
Twan Koolen and Robin Deits.
\newblock Julia for robotics: simulation and real-time control in a high-level
  programming language.
\newblock \emph{2019 International Conference on Robotics and Automation
  (ICRA)}, pages 604--611, 2019.

\bibitem[Kumar(2016)]{Kumar2016thesis}
Vikash Kumar.
\newblock \emph{Manipulators and Manipulation in high dimensional spaces}.
\newblock PhD thesis, University of Washington, Seattle, 2016.
\newblock URL
  \url{https://digital.lib.washington.edu/researchworks/handle/1773/38104}.

\bibitem[Lam et~al.(2015)Lam, Pitrou, and Seibert]{lamNumbaLLVMbasedPython2015}
Siu~Kwan Lam, Antoine Pitrou, and Stanley Seibert.
\newblock Numba: {{A LLVM}}-based {{Python JIT Compiler}}.
\newblock In \emph{Proceedings of the {{Second Workshop}} on the {{LLVM
  Compiler Infrastructure}} in {{HPC}}}, {{LLVM}} '15, pages 7:1--7:6, {New
  York, NY, USA}, 2015. {ACM}.
\newblock ISBN 978-1-4503-4005-2.
\newblock \doi{10.1145/2833157.2833162}.

\bibitem[Lee et~al.(2018)Lee, Grey, Ha, Kunz, Jain, Ye, Srinivasa, Stilman, and
  Liu]{DARTSimulator}
Jeongseok Lee, Michael~X. Grey, Sehoon Ha, Tobias Kunz, Sumit Jain, Yuting Ye,
  Siddhartha~S. Srinivasa, Mike Stilman, and Chuanjian Liu.
\newblock Dart: Dynamic animation and robotics toolkit.
\newblock \emph{J. Open Source Software}, 3:\penalty0 500, 2018.

\bibitem[Lowrey et~al.(2018{\natexlab{a}})Lowrey, Kolev, Dao, Rajeswaran, and
  Todorov]{Lowrey2018ReinforcementLF}
Kendall Lowrey, Svetoslav Kolev, Jeremy Dao, Aravind Rajeswaran, and Emanuel
  Todorov.
\newblock Reinforcement learning for non-prehensile manipulation: Transfer from
  simulation to physical system.
\newblock \emph{CoRR}, abs/1803.10371, 2018{\natexlab{a}}.

\bibitem[Lowrey et~al.(2018{\natexlab{b}})Lowrey, Rajeswaran, Kakade, Todorov,
  and Mordatch]{POLO}
Kendall Lowrey, Aravind Rajeswaran, Sham Kakade, Emanuel Todorov, and Igor
  Mordatch.
\newblock {Plan Online, Learn Offline: Efficient Learning and Exploration via
  Model-Based Control}.
\newblock \emph{ICLR}, abs/1811.01848, 2018{\natexlab{b}}.

\bibitem[Mania et~al.(2018)Mania, Guy, and Recht]{Mania2018SimpleRS}
Horia Mania, Aurelia Guy, and Benjamin Recht.
\newblock Simple random search of static linear policies is competitive for
  reinforcement learning.
\newblock In \emph{NeurIPS}, 2018.

\bibitem[Mordatch et~al.(2015)Mordatch, Lowrey, and
  Todorov]{Mordatch2015EnsembleCIO}
Igor Mordatch, Kendall Lowrey, and Emanuel Todorov.
\newblock Ensemble-cio: Full-body dynamic motion planning that transfers to
  physical humanoids.
\newblock \emph{2015 IEEE/RSJ International Conference on Intelligent Robots
  and Systems (IROS)}, pages 5307--5314, 2015.

\bibitem[Nachum et~al.(2019)Nachum, Ahn, Ponte, Gu, and
  Kumar]{Nachum2019QuadrupedSim2Real}
Ofir Nachum, Michael~J. Ahn, Hugo Ponte, Shixiang Gu, and Vikash Kumar.
\newblock Multi-agent manipulation via locomotion using hierarchical sim2real.
\newblock \emph{ArXiv}, abs/1908.05224, 2019.

\bibitem[Nagabandi et~al.(2019)Nagabandi, Konoglie, Levine, and Kumar]{PDDM}
Anusha Nagabandi, Kurt Konoglie, Sergey Levine, and Vikash Kumar.
\newblock {Deep Dynamics Models for Learning Dexterous Manipulation}.
\newblock In \emph{Conference on Robot Learning (CoRL)}, 2019.

\bibitem[OpenAI(2018)]{OpenAI2018LearningDI}
OpenAI.
\newblock Learning dexterous in-hand manipulation.
\newblock \emph{CoRR}, abs/1808.00177, 2018.

\bibitem[Paszke et~al.(2017)Paszke, Gross, Chintala, Chanan, Yang, Devito, Lin,
  Desmaison, Antiga, and Lerer]{pytorch}
Adam Paszke, Sam Gross, Soumith Chintala, Gregory Chanan, Edward Yang, Zachary
  Devito, Zeming Lin, Alban Desmaison, Luca Antiga, and Adam Lerer.
\newblock Automatic differentiation in pytorch.
\newblock 2017.

\bibitem[Rajeswaran et~al.(2016)Rajeswaran, Ghotra, Ravindran, and
  Levine]{Rajeswaran2016EPOpt}
Aravind Rajeswaran, Sarvjeet Ghotra, Balaraman Ravindran, and Sergey Levine.
\newblock Epopt: Learning robust neural network policies using model ensembles.
\newblock In \emph{ICLR}, 2016.

\bibitem[Rajeswaran et~al.(2017)Rajeswaran, Lowrey, Todorov, and
  Kakade]{Rajeswaran17nips}
Aravind Rajeswaran, Kendall Lowrey, Emanuel Todorov, and Sham Kakade.
\newblock {Towards Generalization and Simplicity in Continuous Control}.
\newblock In \emph{NIPS}, 2017.

\bibitem[Rajeswaran et~al.(2018)Rajeswaran, Kumar, Gupta, Vezzani, Schulman,
  Todorov, and Levine]{Rajeswaran-RSS-18}
Aravind Rajeswaran, Vikash Kumar, Abhishek Gupta, Giulia Vezzani, John
  Schulman, Emanuel Todorov, and Sergey Levine.
\newblock {Learning Complex Dexterous Manipulation with Deep Reinforcement
  Learning and Demonstrations}.
\newblock In \emph{Proceedings of Robotics: Science and Systems (RSS)}, 2018.

\bibitem[Sadeghi and Levine(2016)]{Sadeghi2016CAD2RLRS}
Fereshteh Sadeghi and Sergey Levine.
\newblock Cad2rl: Real single-image flight without a single real image.
\newblock \emph{ArXiv}, abs/1611.04201, 2016.

\bibitem[Schulman et~al.(2015)Schulman, Levine, Moritz, Jordan, and
  Abbeel]{Schulman15}
John Schulman, Sergey Levine, Philipp Moritz, Michael Jordan, and Pieter
  Abbeel.
\newblock Trust region policy optimization.
\newblock In \emph{ICML}, 2015.

\bibitem[Srinivasa et~al.(2009)Srinivasa, Ferguson, Helfrich, Berenson, Collet,
  Diankov, Gallagher, Hollinger, Kuffner, and Weghe]{Srinivasa2009HERBAH}
Siddhartha~S. Srinivasa, Dave Ferguson, Casey Helfrich, Dmitry Berenson, Alvaro
  Collet, Rosen Diankov, Garratt Gallagher, Geoffrey~A. Hollinger, James~J.
  Kuffner, and Michael~Vande Weghe.
\newblock Herb: a home exploring robotic butler.
\newblock \emph{Autonomous Robots}, 28:\penalty0 5--20, 2009.

\bibitem[Tassa et~al.(2018)Tassa, Doron, Muldal, Erez, Li, de~Las~Casas,
  Budden, Abdolmaleki, Merel, Lefrancq, Lillicrap, and
  Riedmiller]{DeepmindDmControl2019}
Yuval Tassa, Yotam Doron, Alistair Muldal, Tom Erez, Yazhe Li, Diego
  de~Las~Casas, David Budden, Abbas Abdolmaleki, Josh Merel, Andrew Lefrancq,
  Timothy Lillicrap, and Martin Riedmiller.
\newblock Deep{Mind} control suite.
\newblock Technical report, DeepMind, January 2018.
\newblock URL \url{https://arxiv.org/abs/1801.00690}.

\bibitem[Tobin et~al.(2017)Tobin, Biewald, Duan, Andrychowicz, Handa, Kumar,
  McGrew, Ray, Schneider, Welinder, Zaremba, and Abbeel]{Tobin2017DRGrasping}
Josh Tobin, Lukas Biewald, Rocky Duan, Marcin Andrychowicz, Ankur Handa, Vikash
  Kumar, Bob McGrew, Alex Ray, Jonas Schneider, Peter Welinder, Wojciech
  Zaremba, and Pieter Abbeel.
\newblock Domain randomization and generative models for robotic grasping.
\newblock \emph{2018 IEEE/RSJ International Conference on Intelligent Robots
  and Systems (IROS)}, pages 3482--3489, 2017.

\bibitem[Todorov et~al.(2012)Todorov, Erez, and Tassa]{mujoco12}
Emanuel Todorov, Tom Erez, and Yuval Tassa.
\newblock Mujoco: A physics engine for model-based control.
\newblock In \emph{IROS}, 2012.

\bibitem[Urakami et~al.(2019)Urakami, Hodgkinson, Carlin, Leu, Rigazio, and
  Abbeel]{Urakami2019DoorGymAS}
Yusuke Urakami, Alec Hodgkinson, Casey Carlin, Randall Leu, Luca Rigazio, and
  Pieter Abbeel.
\newblock Doorgym: A scalable door opening environment and baseline agent.
\newblock \emph{ArXiv}, abs/1908.01887, 2019.

\bibitem[Williams et~al.(2016)Williams, Drews, Goldfain, Rehg, and
  Theodorou]{williams2016aggressive}
Grady Williams, Paul Drews, Brian Goldfain, James~M Rehg, and Evangelos~A
  Theodorou.
\newblock Aggressive driving with model predictive path integral control.
\newblock In \emph{Robotics and Automation (ICRA), 2016 IEEE International
  Conference on}, pages 1433--1440. IEEE, 2016.

\end{thebibliography}

\clearpage
\appendix
\section{AbstractEnvironment Interface}

For a more thorough description of the AbstractEnvironment interface and more, see the documentation at
\url{docs.lyceum.ml/dev}.
\begin{minted}[fontsize=\footnotesize]{julia}
  statespace(env)
  getstate(env)
  getstate!(s, env)
  setstate!(env, s)

  obsspace(env)
  getobs(env)
  getobs!(o, env)

  actionspace(env)
  getaction(env)
  getaction!(a, env)
  setaction!(env, a)

  rewardspace(env)
  getreward(s, a, o, env)

  evalspace(env)          # Task evaluation metric
  geteval(s, a, o, env)   # that can differ from reward

  reset!(env)             # Reset to a fixed initial state.
  randreset!(env)         # Reset to a random initial state.
         
  step!(env)              # Step the environment. 
  isdone(env)             # return `true` if `env` terminated early 
\end{minted}

\section{Example MuJoCo Environment}

The following is an example of a the OpenAI Gym hopper environment ported to \name. Functions that are not defined from the previous section use the default behaviors of the AbstractEnvironment type.

\begin{minted}[fontsize=\footnotesize,
            encoding=utf8,
escapeinside=||]{julia}
# We begin by importing functions from LyceumMuJoCo we wish
# to extend with our environments specifications. We define
# the functions to work with our environment struct.
import LyceumMujoco: getobs!, randreset!, geteval, step!
struct HopperV2 <: AbstractEnvironment   # This thread safe data structure
    sim::MJSim                           # stores the simulator and other
                                         # user-desired values
    function HopperV2()
        new(MJSim("hopper.xml"))
    end
end

function getobs!(o, env::HopperV2)  # writes data to pre-allocated o
    nq = env.sim.m.nq               # 'env' will be the struct above,
    qpos = env.sim.d.qpos[2:end]    # and accessing its fields provides the
    qvel = copy(env.sim.d.qvel)     # desired observations
    clamp!(qvel, -10, 10)
    copyto!(o, vcat(qpos, qvel))
    return o
end

function randreset!(env::HopperV2)  # Reset environment to a random state 
    reset!(env)
    dist = Uniform(-0.005, 0.005)   # Uniform sampler, called multiple times
    env.sim.d.qpos .+= rand.(dist)  # to fill the qpos, qvel vectors
    env.sim.d.qvel .+= rand.(dist)  # '.' notation vectorizes call to `rand`
    forward!(env.sim)               # MuJoCo's forward dynamics
    return env                      # to propagate changes
end

function geteval(s, a, o, env::HopperV2)  # We evaluate distance along x axis
    statespace(env)(s).qpos[1]
end

function getreward(s, a, o, env::HopperV2)
    shapedstate = statespace(env)(s)
    qpos = shapedstate.qpos
    qvel = shapedstate.qvel

    x0 = qpos[1]                                 
    step!(env.sim, a)                           
    x1, height, ang = qpos[1:3]

    alive_bonus = 1.0
    reward = (x0 - x1) / dt(env)
    reward += alive_bonus
    reward -= 1e-3 * sum(x->x^2, a) # lambda function squares `a`
    return reward
end

# # # # # # # # # # # # # # # # # # # # # #
# After creation of the environment we can use it with LyceumAI.
# we first include LyceumAI and other packages.
using LyceumAI, Flux

hop = HopperV2()
dobs, dact = length(obsspace(hop)), length(actionspace(hop))

# A policy and value function are created using helper 
# functions built on Flux.jl
policy = DiagGaussianPolicy(
    multilayer_perceptron(dobs, 64, 64, dact),
    ones(dact) .*= -0.5
)

value = multilayer_perceptron(dobs, 128, 128, 1)
valueloss(bl, X, Y) = mse(vec(bl(X)), vec(Y))

# FluxTrainer is thing you can iterate on. The result at each 
# loop is passed to stopcb below, so you can quit after
# a number of epochs, convergence, both, or never
valuetrainer = FluxTrainer(
    optimiser = ADAM(1e-3),
    szbatch = 64,
    lossfn = valueloss,
    stopcb = s->s.nepochs > 2
)

# The LyceumAI NaturalPolicyGradient is an iterator, where each loop
# the data is returned to the for-loop below. We construct nthreads
# number of Hopper Environments to be parallel evaluated.
npg = NaturalPolicyGradient(
    (HopperV2() for _=1:Threads.nthreads()),
    policy,
    value,
    gamma = 0.995,
    gaelambda = 0.97,
    valuetrainer,
    Hmax=1000,
    norm_step_size=0.1,
    N=10000
)

# We iterate on NPG for 100 iterations, printing useful information
# every 25 iterations.
for (i, state) in enumerate(npg)
    if i > 100
        break
    end
    if mod(i, 25) == 0
        println("stocreward = ", state.stoctraj_reward)
    end
end

\end{minted}

\clearpage
\section{Details on MPC experiments}
For the MPC comparison with MPPI, we considered two environments. A 7-DOF Sawyer robot reaching various spatial goals with the end effector, and a 30-DOF in-hand manipulation task where a Shadow Hand has to manipulate a pen to match a desired orientation. Both tasks are episodic, where at the start of an episode, a random initial configuration and a random target configuration are generated. Each episode is 75 time-steps. The specific MPPI algorithm we used was based on \citet{Lowrey2018ReinforcementLF}, where the authors first observed that correlated noise sequences were important for hand manipulation tasks. Our observations are consistent with this finding. The main hyper-parameters used are summarized in the below table.

\begin{table}[h!]
\begin{center}
\caption{MPC experiment parameters}
\begin{tabular}{|c|c|c|}
\hline
Parameter                   & Shadow Hand experiment        &   Sawyer experiment    \\ \hline
Planning horizon            & 32                            &   16                   \\ \hline
\# trajectories sampled     & 160                           &   30                   \\ \hline
Temperature                 & 1.0                           &   5.0                  \\ \hline
Smoothing parameters        & $\beta_0=0.25$, $\beta_1=0.8$ &   $\beta_0=0.25$, $\beta_1=0.8$  \\ \hline
\end{tabular}
\end{center}
\label{table:mpc_parameters}
\end{table}

\section{Details on RL experiments}
As a representative RL experiment, we use the NPG algorithm and compared \name~with Gym. We study the learning curve as a function of both the number of environment interactions as well as wall clock time. As a function of environment interactions, we found Gym and \name~to be comparable (as expected), however \name~was substantially faster in wall clock time. The hyper-parameters for the RL experiment are mentioned in Table~\ref{table:npg_parameters}.

\begin{table}[h!]
\caption{Parameters for the RL (NPG) experiment}
\begin{center}
\begin{tabular}{|l|l|}
\hline
Parameter               & Value         \\ \hline
\# NPG steps            & 100           \\ \hline
Samples per NPG step    & 10,000        \\ \hline
NPG step size (normalized)  & 0.1        \\ \hline
Policy size             & (64, 64)      \\ \hline
Value function size     & (128, 128)    \\ \hline
Discount $(\gamma)$     & $0.995$        \\ \hline
GAE $(\lambda)$         & $0.97$       \\ \hline
\end{tabular}
\end{center}
\label{table:npg_parameters}
\end{table}

\end{document}